**VLM6D: RGB-D 영상을 활용한 비전 네트워크 기반의 6자유도 물체 자세 추정 기법**

## VLM6D: VLM based 6Dof Pose Estimation based on RGB-D Images

**Md Selim Sarowar**   **Sungho Kim**

Selim.sarowar12@gmail.com, sunghokim@yu.ac.kr
School of Electronics Engineering,
Advanced Visual Intelligence Lab(AVI),
Yeungnam University, **South Korea**

### Abstract

The primary challenge in computer vision is precisely calculating the pose of 6D objects, however many current approaches are still fragile and have trouble generalizing from synthetic data to real-world situations with fluctuating lighting, textureless objects, and significant occlusions. To address these limitations, VLM6D, a novel dual-stream architecture that leverages the distinct strengths of visual and geometric data from RGB-D input for robust and precise pose estimation. Our framework uniquely integrates two specialized encoders: a powerful, self-supervised Vision Transformer (DINOv2) processes the RGB modality, harnessing its rich, pre-trained understanding of visual grammar to achieve remarkable resilience against texture and lighting variations. Concurrently, a PointNet++ encoder processes the 3D point cloud derived from depth data, enabling robust geometric reasoning that excels even with the sparse, fragmented data typical of severe occlusion. These complementary feature streams are effectively fused to inform a multi-task prediction head. We demonstrate through comprehensive experiments that VLM6D obtained new SOTA performance on the challenging Occluded-LineMOD, validating its superior robustness and accuracy.

### 1. Introduction

In the intricate dance between a robot and its world, the ability to perceive and understand the orientation of objects is paramount. This task, known as 6D object pose estimation, is the silent language that enables a machine to grasp, manipulate, and interact with its environment. From the precise ballet of an automated assembly line to the immersive overlays of augmented reality, accurately determining an object's position and rotation from sensor data is a foundational challenge in modern computer vision.

Despite all of the advancements made by deep learning, many modern techniques are still fragile. They frequently struggle in the unpredictable turmoil of the actual world because they were trained in the sterile, predictable bounds of synthetic datasets. Their perception can be broken by the brief brightness of a light, the shifting shadows of a cluttered shelf, or the glimpse of an object peering from behind another[3]. This is especially true for models that have already been pre-trained under supervision on particular domains; their acquired knowledge is frequently too inflexible and does not generalize when faced with new textures, lighting, or the extreme occlusions that characterize difficult benchmarks such as the LineMOD-Occluded (LMO) dataset[15].

In this work, we proposed a novel approach to 6D pose estimation that is based on extensive generality and inherent robustness. Two specialized encoders, each of which is an expert in a distinct sensory modality, make up our architecture[5]. Instead of using a traditionally trained network for the visual world of color, we use DINOv2, a potent, self-supervised Vision Transformer. DINOv2, which has been pre-trained on a massive dataset of 142 million unlabeled photos, has acquired a basic visual grammar. It has a deep understanding of texture, form, and context that enables it to accomplish amazing zero-shot transfer feats. It also exhibits a surprising resilience to the very fluctuations in lighting and texture that lead to the failure of inferior models. However, the geometric truth is retained even when occlusion steals the visual tale. Our second stream uses PointNet++, an architecture that can reason directly in three dimensions, to do this. PointNet++ traverses the sparse, fragmented archipelago of a partial point cloud, whereas RGB-based techniques simply observe a void. It is particularly well-suited to the challenge of the LMO collection, where objects can be up to 80% obscured, because of its hierarchical structure, which enables it to sense both the fine-grained features and the overall shape of an object. It may find geometric certainty in the middle of visual confusion because of its intrinsic permutation invariance, which renders it immune to the disorder of jumbled points.

## 2. Related Work

### 2.1 Overview

A core problem in 3D computer vision, estimating an object's 6D position has important implications in autonomous driving, and robotic manipulation. From techniques that relied on individual RGB images to those that made use of the rich geometric information offered by RGB-D sensors, research in this field has undergone a substantial transformation[7][8][9]. Despite their shown effectiveness, RGB-only techniques are intrinsically constrained when handling objects without texture or in areas with extreme occlusions, which might result in inaccurate 2D–3D correspondences[10]. Two prominent deep learning paradigms—direct pose prediction and correspondence-based prediction—have emerged as a result of the growing availability of depth sensors, which has caused high-precision pose estimation to move toward RGB-D data.

### 2.2 Intuitive Pose Prediction

Methods in this category learn a direct mapping between fused RGB-D features and the final 6D posture parameters. These approaches are frequently preferred for their computational efficiency and end-to-end simplicity[10][11]. DenseFusion, a pioneering effort, introduced a per-pixel fusion technique for combining color and depth variables prior to position regression. Subsequent strategies attempted to improve on this foundation. For example, ES6D introduced XYZNet, a fully convolutional architecture, to improve computational efficiency, as well as a symmetry-invariant loss function to handle symmetric objects more effectively[13]. Uni6Dv2 has recently concentrated on enhancing robustness by introducing a two-step pipeline to reduce noise from both background pixels ("Instance-Outside") and incorrect depth measurements ("Instance-Inside") prior to final regression[14].

### 2.3 Residual-Based Prediction

In this context, the reference for our work, the Residual-based Dense Point-wise Network (RDPN6D),[10] likewise uses dense correspondence to forecast the 3D model coordinates for each visible pixel. Its main innovation is its residual representation, which turns the unsolvable problem of regressing absolute 3D coordinates into a more stable combination of a fine-grained, bounded regression task (predicting the small offset vector from that anchor) and a coarse classification task (predicting the closest "anchor" region on the object)[10]. These dense techniques deliver state-of-the-art performance and exhibit greater resilience, especially in scenarios with significant occlusion, by utilizing information from the full visible surface.

The potential of dense correspondence for reliable posture estimation is validated by RDPN6D's success. However, it might not be as reliable as conventional geometric solvers because it relies on a learnt CNN to regress the final pose from the dense coordinate map. Additionally, a large computational and memory footprint is introduced by anticipating a dense map for every pixel as well as unconsolitated network. In this study, we fill these gaps and expand on the fundamental ideas of dense correspondence[12][13]. We suggest a new architecture that achieves a better mix of accuracy, robustness, and real-time performance.

## 3. Methodology

Our proposed architecture follows a dual-stream design that processes RGB and depth modalities independently before fusing them for pose prediction. The pipeline consists of four main components: (1) RGB feature extraction using self-supervised DINOv2, (2) depth-to-point cloud conversion and geometric feature learning via PointNet++, (3) cross-modal feature fusion, and (4) multi-task prediction heads. Figure 2 illustrates the complete architecture.

The input RGB image $I_{RGB} \in \mathbb{R}^{H \times W \times 3}$ is first resized to 224×224 & normalized using ImageNet statistics.[6] Simultaneously, the depth image $I_D \in \mathbb{R}^{H \times W}$ is back-projected to 3D space using camera intrinsics to generate a colored point cloud. The dual-stream encoders produce feature vectors $f_{RGB} \in R^{768}$ and $f_{depth} \in R^{1024}$, which are concatenated and processed through a fusion network to obtain a shared representation $f_{fused} \in R^{512}$ for final pose prediction.

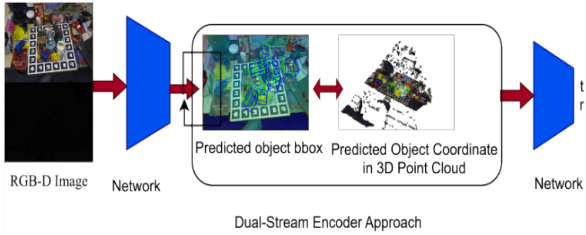

Figure 1: **Overview of VLM6D (Vision Language Model 6 Degree of Freedom) approach.** Our model follows a **two-stream fusion approach** that processes RGB and Depth data separately before combining them.

3.1 RGB Feature Extraction with DINOv2

We employ DINOv2 (ViT-B/14) as our RGB encoder due to its superior performance in learning robust visual representations through self-supervised learning. DINOv2 is the ViT pre-trained containing weights, biases and on 142 million curated images using self-distillation with no labels (DINO), which provides strong generalization capabilities across diverse visual domains.

The input image is splitted into non-overlapping patches of size 16×16 pixels, resulting in $N_{patches}$=224/16×224/16=196 patches. Each patch is linearly projected to a 768-dimensional embedding.

**Transformer Encoder:** The patch embeddings are prepended with positional encodings and a learnable class token [CLS]. The resulting sequence passes through 12 transformer layers, each comprising of feed-forward network (FFN) and multi-head self-attention (MHSA).

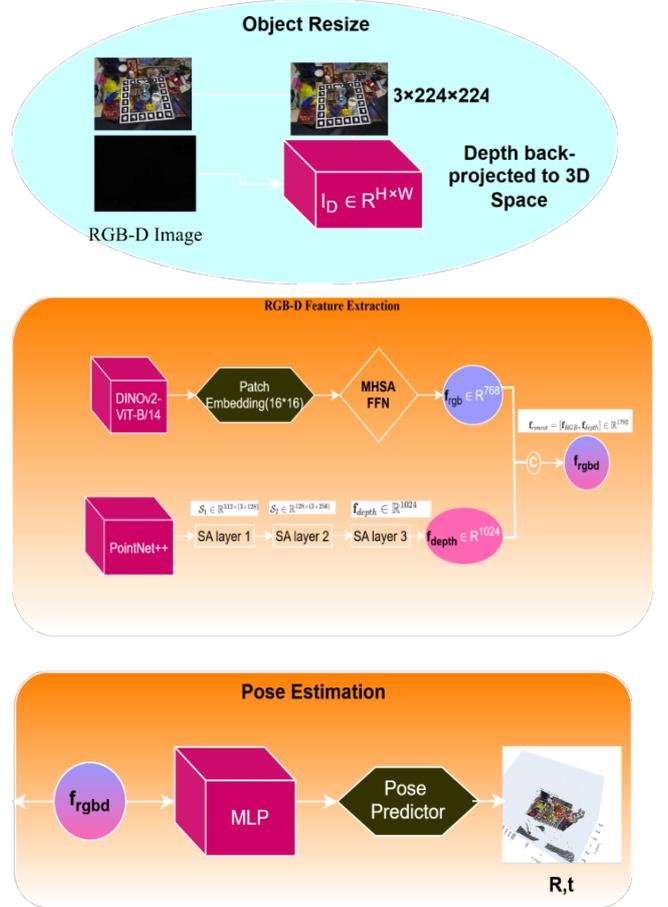

Figure 2: **Framework of VLM6D.** i) starting with an RGB-D image to resize it and Depth data processing to 3D space. ii) Once we have prepared the data, ViT based RGB encoder(DINOv2) will take it and PointNet++ as depth data. After feature extractions and processing, It will concate and passing through Cross Modal Feature Fusion. iii) from the fused features, predicted the outputs through separate prediction heads.

3.2 Depth Processing with PointNet++

PointNet++ hierarchically learns geometric features from the point cloud through multiple Set Abstraction (SA) layers. Each SA layer consists of three operations: sampling, grouping, and local feature aggregation.

Layer 1 - Local Feature Extraction:

Sampling: Apply FPS to select $N_1$=512 representative points from the input N=2048.

Feature Learning: Process each local neighborhood through PointNet (shared MLPs: 3→64→64→128) followed by max pooling.

Layer 2 - Regional Feature Aggregation:
- Sampling: FPS to select $N_2=128$ $S_1$
- Feature Learning: MLPs: (3+128)→128→128→256

Layer 3 - Global Feature Extraction:
- Global Aggregation: All 128 points in $S_2$ are processed together
- Feature Learning: MLPs: (3+256)→256→512→1024
- Max Pooling: Aggregate across all points to obtain global descriptor

3.3 Cross-Modal Feature Fusion

The RGB and depth features are complementary: RGB provides appearance, texture, and semantic information, while depth captures geometric structure and 3D shape. We fuse these modalities through late fusion by concatenating the feature vectors:

$$f_{concat}=[f_{RGB},f_{depth}]\in R^{1792} \quad \ldots\ldots(1)$$

The concatenated features are endured processed through a two-layer MLP with ReLU activations and dropout for regularization:

$$h_1=Dropout(ReLU(Linear_{1792\to1024}(f_{concat}))) \quad \ldots\ldots(2)$$
$$f_{fused}= Dropout(ReLU(Linear1_{024\to512}(h_1))) \quad \ldots\ldots(3)$$

We use a dropout rate of 0.3 to prevent overfitting. This fusion strategy allows each modality to be processed with its optimal architecture while learning joint representations for pose estimation.

3.4 Multi-Task Prediction Heads

From the fused features $f_{fused}$, we predict four outputs through separate prediction heads:
1) Rotation Prediction
2) Translation Prediction
3) Confidence Prediction
4) Object Classification

### 4. Experiments

This section includes comparisons to the other methods as well as the experimental findings of our VLM6D method.

Table 1: Experimental results of 6D objects pose metric of ADD(-S) on LM-O dataset.

| Method | PoseCNN | Hybridpose | PVN3D [17] | FFB6D [16] | RCVPose [42] | Uni6D [23] | Uni6Dv2 [38] | DFTr [46] | RDPN | VLM6D(Ours) |
|---|---|---|---|---|---|---|---|---|---|---|
| ape | 9.6 | 20.9 | 33.9 | 47.2 | 60.3 | 33 | 44.3 | 64.1 | 64.6 | **81** |
| can | 45.2 | 75.3 | 88.6 | 85.2 | 92.5 | 51 | 53.3 | 96.1 | 97 | 78.9 |
| cat | 0.9 | 24.9 | 39.1 | 45.7 | 50.2 | 4.6 | 16.7 | 52.2 | 54.8 | **86.7** |
| driller | 41.4 | 70.2 | 78.4 | 81.4 | 78.2 | 58.4 | 63 | 95.8 | 93.1 | 81.9 |
| duck | 19.6 | 27.9 | 41.9 | 53.9 | 52.1 | 34.8 | 38.1 | 72.3 | 68.8 | **75** |
| eggbox | 22 | 52.4 | 80.9 | 70.2 | 81.2 | 1.7 | 4.6 | 75.3 | 78.1 | **83.4** |
| glue | 38.5 | 53.8 | 68.1 | 60.1 | 72.1 | 30.2 | 40.3 | 79.3 | 83.5 | 79 |
| holepuncher | 22.1 | 54.2 | 74.7 | 85.9 | 75.2 | 32.1 | 50.9 | 86.8 | 96.1 | 81 |
| Avg (8) | 24.9 | 47.5 | 63.2 | 66.2 | 70.2 | 30.7 | 40.2 | 77.7 | 79.5 | 81.6 |

The LineMOD (LMO) dataset, a well-known and difficult benchmark for 6D pose estimation, served as the basis for our investigations. Thirteen items in cluttered environments with issues like minor occlusion, fluctuating lighting, and texture-less object surfaces make up the main LineMOD dataset[3][15]. We adhered to the accepted evaluation procedure set forth in earlier studies. We also conducted tests on the LMO dataset to evaluate our architecture resilience to large occlusions. Using the original LineMOD data, this addon offers difficult test cases with severely occluded objects.

In accordance with earlier research, we employ the widely used average distance metrics ADD(-S).

**Comparison with SOTA Methods:**

The effectiveness of VLM6D was demonstrated on the RDPN (Tab. 1), where it reaffirms its position as the state-of-the-art. It sets a new standard for robustness in addition to surpassing current approaches on the ADD-(S) metric, demonstrating its strong generalizability. In challenging situations where modern RGB-D solutions usually fail, such as extreme occlusions and objects that are textureless or very reflective, VLM6D continues to function quite well.

### Conclusion

VLM6D, a novel & robust architecture for 6D (translation & rotation) object pose estimation based on RGB-D data. By obtaining outstanding results on the ADD-(S) measure, it demonstrates its exceptional precision and applicability. More significantly, VLM6D exhibits remarkable resilience against reflecting surfaces, texture-less objects, and extreme occlusions—conditions that are infamously challenging for RGB-D-based techniques.

**Acknowledgement**

This research was supported by the Gyeongsangbuk-do RISE(Regional Innovation System & Education) project.